\documentclass[10pt,conference]{IEEEtran}
\IEEEoverridecommandlockouts
\usepackage{cite}
\usepackage{amsmath,amssymb,amsfonts,commath}
\usepackage{algorithmic}
\usepackage{algorithm}

\usepackage{graphicx}
\usepackage{textcomp}
\usepackage{xcolor}

\usepackage{amsthm}
\usepackage{regexpatch,fancyvrb,xparse}

\usepackage{mathtools}
\usepackage{color}
\usepackage{soul}

\def\BibTeX{{\rm B\kern-.05em{\sc i\kern-.025em b}\kern-.08em
    T\kern-.1667em\lower.7ex\hbox{E}\kern-.125emX}}

\begin{document}

\title{Exploring Vision Neural Network Pruning via Screening Methodology}

 \author{
 \IEEEauthorblockN{Mingyuan Wang}
 \IEEEauthorblockA{\textit{Department of Statistics} \\
 \textit{Florida State University}\\
 Tallahassee, Florida, USA \\
 bruce.wmy.research@gmail.com}
 \and
 \IEEEauthorblockN{Yangzi Guo}
 \IEEEauthorblockA{\textit{GPU ML Accelerator Team} \\
 \textit{Qualcomm Inc.}\\
 San Diego, California, USA \\
 yangguo@qti.qualcomm.com}
 \and
 \IEEEauthorblockN{Sida Liu}
 \IEEEauthorblockA{\textit{Department of Statistics} \\
 \textit{Florida State University}\\
 Tallahassee, Florida, USA \\
 lsd930808@gmail.com}
 \and
 \IEEEauthorblockN{Yuhang Liu}
 \IEEEauthorblockA{\textit{Department of Statistics} \\
 \textit{Florida State University}\\
 Tallahassee, Florida, USA \\
 yuhang.stat@gmail.com}
 }

\maketitle

\begin{abstract}
The remarkable performance of modern deep neural networks (DNNs) is largely driven by their massive scale, often comprising tens to hundreds of millions—or even billions—of parameters. However, such a scale incurs substantial storage and computational costs, hindering deployment on platforms such as edge devices that require energy-efficient and real-time processing. In this paper, we propose a network pruning framework that reduces both storage and computation requirements by an order of magnitude while preserving model accuracy. Our approach eliminates non-essential parameters through a statistical analysis of component significance across classification categories. Specifically, we employ a F-statistic-based screening technique combined with a weighted evaluation scheme to quantify the contributions of connections and channels, enabling both unstructured and structured pruning within a unified framework. Extensive experiments on real-world vision datasets, covering both fully connected neural networks (FNNs) and convolutional neural networks (CNNs), demonstrate that the proposed framework produces compact and efficient models that are highly competitive with the state of art apporoaches. %\hl{Should we add that we will share the code after acceptance?}
\end{abstract}

\section{Introduction}
Deep Neural Networks or Deep Learning have revolutionized numerous fields, achieving remarkable success in computer vision \cite{krizhevsky2012imagenet, he2017mask}, natural language processing \cite{mikolov2013efficient, sutskever2014sequence, devlin2018bert}, predictive analytics \cite {guo2017deepfm, arora2022probabilistic, lim2021temporal}, and beyond. Despite these advances, the large computational and memory requirements of such models challenge real-time applications and resource-constrained device deployments, such as mobile phones and edge computing platforms. These environments require lightweight models that can balance fast inference speed and acceptable performance, underlined by an urgent need for effective model compression techniques.

These challenges have further spawned various approaches to optimize deep neural networks, such as network pruning \cite{cheng2024survey, han2015learning, liu2017learning}, knowledge distillation \cite {xu2024survey, hinton2015distilling, tian2019contrastive}, weight quantization \cite{lin2024awq, jacob2018quantization, wu2018training} and neural sparsity encoding \cite{wang2023co, wu2018deep, dawer2020neural}, to name a few. Of these, network pruning stands out as one of the effective model compression techniques through the systematic removal of redundant elements like weights, filters, or channels within models. Pruning reduces computational complexity and savings in storage while increasing the potential for inference speed; hence, it has been one practical mean of deploying deep learning models on resource-constrained devices. Loosely speaking, pruning techniques could be divided into unstructured \cite{vadera2022methods} and structured \cite{lemaire2019structured}. 

The network pruning process typically begins with an either trained or untrained DNN, and two main strategies which are iterative pruning and one-time pruning can be employed. Iterative removal procedure is widely applied not only in neural network pruning fields but also in fields like feature selection \cite{Barbu_undated}, graph network optimization \cite{Yang2021-iq}, and beyond. Iterative pruning \cite{pmlr-v119-tan20a, Guo9533741} gradually removes unimportant elements over multiple steps, often guided by a decay or annealing function, allowing the network to adapt and recover through retraining after each step. In contrast, one-time pruning \cite{liu2017learning, Amersfoort2020SingleSS} eliminates all unimportant components in a single step, offering faster execution but requiring careful tuning to minimize performance degradation. A critical aspect of pruning is selecting an appropriate metric to identify elements for removal. The choice of metric is closely tied to the pruning granularity—weight pruning focuses on fine-grained optimization by removing individual weights, while channel pruning emphasizes structural sparsity by eliminating entire filters or channels.

In this paper, our major contributions can be summarized as follows:
\begin{itemize}
    \item We propose a novel network pruning framework that leverages statistical analysis to quantify the significance of network components across classification categories, enabling efficient identification and removal of non-essential parameters.
    \item Our approach systematically evaluates the contributions of weights/connections and channels in deep neural networks, enabling unified unstructured and structured pruning. Specifically, we introduce an F-statistic–based screening mechanism with a weighted evaluation scheme to quantify the importance of network components (e.g., weights/connections and channels).
    \item Extensive experiments on benchmark vision datasets, including MNIST, CIFAR, and SVHN, demonstrate the effectiveness of the proposed method. The results underscore the robustness of the screening mechanism and its strong complementarity with existing pruning approaches.
\end{itemize}

\section{Related Work}
%\subsection{Deep Neural Network Pruning} 
%The field of deep neural network pruning has made significant progress, with many studies and surveys providing detailed overviews of the techniques and their applications. For example, \cite{cheng2024survey} organizes pruning methods based on factors such as when and how pruning is applied, as well as their integration with other compression techniques. This survey also discusses recent developments, such as pruning for large language models and vision transformers, and provides insights into the trade-offs involved and future research directions. Similarly,  \cite{he2023structured} focuses on structured pruning methods, which aim to remove entire filters or channels to achieve practical acceleration. These methods are more hardware-efficient compared to unstructured pruning. The survey highlights key techniques such as filter ranking and dynamic execution strategies while comparing structured and unstructured approaches.

\subsection{Unstructured Weight Pruning} 
Unstructured weight pruning is aimed at reducing the number of parameters in a neural network by removing unimportant or redundant weights or connections. One of the simplest and most efficient strategies for pruning weights relies on the magnitudes \cite{han2015learning, Zhu2017ToPO, Guo9533741, Guo9533833}. This assumes that weights of smaller magnitudes are less important and thus contribute less to the performance of a model. It is a computationally inexpensive approach, because one needs only the computation of either $l_1$ or $l_2$ norms, which, in turn, can be highly scalable for big models. On that aspect, \cite{guo2016dynamic} developed a pruning strategy that integrates the aspect of connection slicing during pruning to evade incorrect pruning. Connection slicing allows the pruning of parameters in whole groups or slabs and ensures the structural integrity of the network. Such methods thus tend to avoid removing key connections that are crucial for a model's good performance. \cite{Lin2020DynamicMP} and \cite{Savarese2019WinningTL} leverage gradient-based criteria for unstructured weight pruning by refining the pruning process by modifying aspects such as the behavior of forward propagation or approximating $l_0$ regularization. \cite{shi2024towards} formulates the core unstructured pruning objective using an energy consumption model derived from Synaptic Operations (SOPs).

\subsection{Structured Channel Pruning} 
Structured channel pruning \cite{gao2018dynamic, chen2020storage, gao2024bilevelpruning, ning2020dsa, liu2022evolving, pei2022neural} offers much more significant acceleration of the model. The primary focus of structured channel pruning lies in removing whole channels, providing considerable gain during deployment, especially in hardware environments. 
There are methods to do structure pruning by selecting the trainable channels. \cite{Luo2018AutoPrunerAE} select the target channels based on the importance score of neurons. \cite{liu2017learning} conduct channel pruning with regularization by scaling factors over Batchnorm (BN) layer. \cite{Ding2019CentripetalSF} generates target filters/channels for slimming down models, while \cite{Zhang2021CarryingOC} chooses the channels contributing more to the network output. 

Recent advancements further introduce diverse strategies for evaluating channel importance. 
Kullback-Leibler (KL) divergence-based methods \cite{Luo2019NeuralNP} quantify the channel information loss between the original and pruned networks by measuring the statistical distance between their output distributions.
Simulated annealing-based approaches \cite{Nayman2019XNASNA} adapt the classical annealing-based optimization algorithm to the channel pruning context. Importance sampling techniques \cite{Baykal2019SiPPingNN} leverage statistical sampling theory to estimate channel importance efficiently.

%Some of the channel pruning methods have been found in few studies which have demonstrated the importance of convolutional filters. For example, [12] measure the sum of absolute weights to decide the importance of filters. [16] compute the percentage of zero activation after ReLU and prune those with a high percentage of zeros. While [17] have proposed an iterative two-step method for channel selection by using LASSO regression and least square reconstruction. Furthermore, the L1-norm constraint has been enforced on Batch Normalization layers in [13] to remove filters with smaller values.

%Other improvements also include methods that add additional loss terms during training for the purpose of strengthening pruning. For instance, [18] enforce a clustering loss to make the filters in a cluster similar and prune those similar filters. [19] propose a greedy algorithm to choose channels layer by layer by constructing a certain optimization problem, providing another efficient mechanism of structured pruning.

%These vary from the most straightforward, norm-based methods to complex, data-driven, and training-aware approaches, therefore illustrating the diversity in the area of structured pruning methods. This form of pruning targets the whole filters or channels, can achieve considerable model acceleration with very minimal performance degradation, and is, hence, a very effective tool in the optimization of neural networks for practical applications.

% sun2024novel,
\subsection{Screening Methodologies} %Feature selection has long been a cornerstone of machine learning, as it reduces model complexity, improves generalization performance, and mitigates overfitting \cite{yin2023igrf, kishor2024early}. It also serves as an important inspiration for score-based pruning methods. A comprehensive survey by \cite{dhal2022comprehensive} categorizes feature selection techniques into screening, wrapper, and embedded methods, highlighting their applications across diverse domains. Among these, screening methods have gained significant attention due to their computational efficiency and strong adaptability to high-dimensional data. In particular, \cite{wang2019screening} provides an extensive empirical evaluation of screening methods, demonstrating their effectiveness in both regression and classification tasks. 
In the feature selection literature, screening methods are filter-based techniques that evaluate each feature independently (or with minimal interaction) using statistical criteria, retaining only the most relevant subset of features. Due to their computational efficiency and scalability, screening methods have been widely adopted across various machine learning applications. For instance, \cite{zhou2022feature} proposes a mutual information–based screening approach (MI-SIS), which demonstrates improved classification performance on patient voice data.
Screening methods have also been extended to neural networks, particularly in text mining tasks. For example, \cite{wang2021feature} introduces a $\chi^2$-based feature selection method to eliminate redundant textual features and enhance model efficiency. More recently, variants of traditional screening techniques have been developed to address challenges in modern data settings. Notably, \cite{wang2022online} presents a family of online screening methods designed for large-scale streaming data with sparsity and concept drift, further expanding the applicability of screening approaches.

% \textcolor{red}{need add more here.} 
% Recently more variants of the traditional screening methods were introduced to handle the newer challenges in modern data. a set of online screening methods were developed in \cite{wang2022online} for large streaming data with sparsity and concept drifting property. 

\section{Proposed Method}
\label{sec:pagestyle}

\subsection{Network Pruning Problem Formulation} \label{sec:screeningM}
Given a dataset $\{(\mathbf{x}^i, y^i)\}_{i=1}^{N}$, where $\mathbf{x}^i$ represents input features and $y^i$ is the corresponding target output, we aim to address the neural network pruning problem using constrained optimization. Let the parameters of the network be denoted by $\mathcal{W} = \{(\mathbf{W_j}, \mathbf{b_j})_{j=1}^{L}\}$ where $\mathbf{W_j}$ and $\mathbf{b_j}$ represent the weight matrix and bias vector in each network layer. Let the loss function be denoted as $L(\mathcal{W})$. The non-structured weight pruning task can be expressed as
\begin{alignat}{1}
\min_{\mathcal{W}} \quad & L(\mathcal{W}) \ \ \ \ \ \ \ \ \ \ \ \mbox{s.t.}\quad ||\mathcal{W^{*}}||_0 \leq r||\mathcal{W}||_0
\label{cons_weight}
\end{alignat}
where the $L_0$ norm restricts  the number of non-zero parameters $||\mathcal{W^{*}}||_0$  to a fraction $r$ of $||\mathcal{W}||_0$, and $r$ denotes a positive float in range $(0, 1)$. In this approach, the individual parameters in the parameter space $\mathcal{W}$ are eliminated irrespective of the position, which ultimately leads to sparse weights irregularly across layers.

When pruning focuses on structured elements of CNNs, such as channels, we can reformulate the problem as
\begin{alignat}{2}
\min_{\mathcal{W}} \quad & L(\mathcal{W}) \ \ \ \ \ \ \ \ \ \ \ \mbox{s.t.}\quad ||\mathcal{C^{*}}||_0 \leq r||\mathcal{C}||_0
\label{cons_channel}
\end{alignat}
where $\mathcal{C} = \{ c_1, \dots, c_M \}$ is the set of channels within the network, and fraction $r$ bounds the number of non-zero channels to $||\mathcal{C^{*}}||_0$. It therefore yields structured sparsity where zero parameters are localized within particular channels. This makes the approach more hardware-friendly and thus much easier to realize in real-world systems.

In this work, we consider the study of weight-level pruning for general networks and channel-level pruning for CNNs that involve Batch Normalization layers as a representative family of models where structured sparsity is induced due to these pruning methods. Solving these formulations would allow us to enhance computational efficiency while preserving predictive model performance.

%In this section, the procedure and mechanism of the Network Pruning with F-Statistic Screening framework are described in detail. First, the classical network pruning problem is formally introduced and mathematically formulated. Then the screening methods F-statistic score will be deduced and explained how it is good metric help assess the importance of weights, connections or channels to categorical classes. After that, we show the implementation of pruning procedure in our framework for two different types of neural network pruning, i.e. unstructured weight pruning and structured channel pruning..

\subsection{From F-Statistic-Based Screening to Network Pruning} \label{sec:screeningM}
Among various screening methodologies, we conducted extensive explorations; however, due to space limitations, this paper focuses on F-statistic–based screening to investigate its applicability to network pruning. Our experimental results demonstrate that F-statistic–based screening effectively balances computational efficiency and model accuracy.

\textbf{F-statistic Screening Score} The classical F-statistic–based screening methodology evaluates feature importance by computing the F-statistic score. %\hl{The F-statistic captures feature importance by measuring class separability. Informative features exhibit large differences in class-wise means and low intra-class variance, leading to high F-statistic scores. In contrast, uninformative features show similar distributions across classes or high intra-class variability, resulting in low F-statistic scores.}

Consider a classification dataset $\{(\mathbf{x}^i, y^i)\}_{i=1}^{N}$, where $N$ is the number of samples in training set, $\mathbf{x}^i \in \mathbb{R}^p$ denotes a sample with $p$-dimensional features, and $y^i \in \{{1, \dots, K}\}$ represents the corresponding class label among $K$ categories. For each feature $x_j$, the F-statistic score is computed by first partitioning the samples into $K$ groups according to their class labels, and then evaluating the feature importance of $x_j$ by gauging the deviation of the group means from the population mean adjusted by variances. In essence, the F-statistic for feature $x_j$ is defined as:
\begin{alignat}{2}
F(x_j)=\frac{\frac{N_k}{K-1}\sum_{k=1}^{K}(\overline{x_{jk}} - \overline{x_j})^2}{\frac{1}{N-K}\sum_{k=1}^{K} \sum_{x_j | y \in k} (x_{j} - \overline{{x_{jk}}})^2}
\label{eq:F}
\end{alignat}
where $N_k$ denotes the number of samples in class $k$, and $N$ is the total number of samples. $\overline{x}_{jk}$ represents the mean value of feature $x_j$ within class $k$, while $\overline{x}_j$ denotes the overall mean of feature $x_j$ across all samples. For a given dataset, an F-statistic score can be computed for each feature.

%\hl{This paragraph duplicates the previous highlighted sentences. suggest to delete entirely.} 
The F-statistic captures feature importance by measuring class separability. Informative features induce large inter-class differences while maintaining low intra-class variability, resulting in high F-statistic scores and facilitating reliable discrimination. This property is directly quantified by the F-statistic as the ratio of between-class variance to within-class variance. A higher F-statistic indicates stronger discriminative capability, as the feature produces well-separated class distributions with minimal intra-class noise. In contrast, uninformative features exhibit similar distributions across classes or high intra-class variability, leading to low F-statistic scores.

\textbf{F-statistic Screening on Network Pruning}
%\hl{To calculate F-statistics for each weight or channel, we generate the weight/channel feature vector $\{(p_i)\}_{i=1}^{N}$. Specifically for F-statistics of weight $w$, given it is in layer $l$.} 
%$$
%p_i = L_{lwi} \cdot w_i
%\label{feature_vector}
%$$
%\hl{Where $L_{lwi}$ is the layer input of layer $l$ that corresponds to weight $w$ after forward passing of sample $i$, and $w_i$ is the weight value at that time. For F-statistics of channel $c$, given it is in layer $l$.}
%$$
%p_i = L_{lci} \cdot w_{ci}
%\label{feature_vector}
%$$
%\hl{Where $L_{lci}$ is the layer input of layer $l$ that corresponds to channel $c$ after forward passing of sample $i$, and $w_{ci}$ is the weight values corresponding to channel $c$ at that time. This yields a feature vector $\{(p_i)\}_{i=1}^{N}$ of the same size as the class-label vector $\{(y^i)\}_{i=1}^{N}$ for each weight/channel. Using $\{(p_i)\}_{i=1}^{N}$ as $x_j$ in Equation} \ref{eq:F}, \hl{we obtain the F-statistics of weight/channel.}
Traditional neural network pruning methods typically do not explicitly account for the influence of training samples during the pruning process. In contrast, the F-statistic provides a principled approach for screening the importance of network components—such as weights and channels—by measuring their discriminative power with respect to class labels. Motivated by this observation, we define a unified ranking metric that integrates traditional magnitude-based pruning with an F-statistic–based screening mechanism. The F-statistic component captures the statistical significance of each network component, while the magnitude term reflects its numerical contribution to the model. These two components are combined in a weighted manner to form the final ranking score, enabling more informed and effective pruning decisions.

In the case of unstructured weight pruning, we leverage the F-statistic score $F(w)$ to evaluate the significance of network weights across classification categories and combine it with weight magnitude to define the ranking metric $\mathcal{M}$. During training, each input sample contributes to parameter updates via backpropagation, allowing each weight $w$ to be associated with a set of values $\{w_i\}_{i=1}^{N}$ over an epoch. For mini-batch training, since weights are shared across samples within a batch, we approximate sample-specific contributions by multiplying each weight with its corresponding input feature values, yielding distinct per-sample representations.
\begin{equation}
\begin{aligned}
\mathcal{M}(w) = \alpha \cdot F(w) + (1 - \alpha) \cdot |w|, w \in \mathcal{W}, \alpha \in(0, 1]
\label{M_unstructured}
\end{aligned}
\end{equation}

In the case of structured channel pruning, we leverage the scale parameters of Batch Normalization (BN) layers following convolutional layers, as BN has become a standard component in modern CNNs due to its effectiveness in accelerating training and improving convergence. The transformation performed by a BN layer is defined as follows:
$$
BN(z_{in}) = \frac{z_{in}-\mu_\mathbf{B}}{\sqrt{\sigma^{2}_{\mathbf{B}}+\epsilon}}; z_{out}=\gamma\cdot BN(z_{in}) + \beta
\label{BN_scale}
$$
where $\mu_{\mathbf{B}}$ and $\sigma_{\mathbf{B}}$ denote the mean and variance computed over the mini-batch $\mathbf{B}$, and $\gamma$ and $\beta$ are trainable scale and shift parameters, respectively. Notably, in Batch Normalization, each input channel (i.e., the output channel of the preceding convolutional layer) is associated with an individual scaling parameter $\gamma_c$, which modulates the contribution of that channel and thus reflects its importance. Building on this, we incorporate the F-statistic score $F(\gamma_c)$ to evaluate the discriminative significance of network channels across classification categories. By combining the $L_1$-norm of $\gamma$ \cite{liu2017learning} with the proposed screening score, we define a unified ranking metric $\mathcal{M}$ for channel pruning.
\begin{equation}
\begin{aligned}
 \mathcal{M}(c) = \alpha \cdot F(\gamma_c) + (1 - \alpha) \cdot |\gamma_c|, c \in \mathcal{C}, \alpha \in(0, 1]
\end{aligned}
\label{eq:M_structured}
\end{equation}

\subsection{Algorithm Description}\label{sec:pruneM}
Our algorithm is built upon three key principles: (1) a unified ranking metric $\mathcal{M}$ that integrates F-statistic–based screening with parameter magnitude to evaluate the importance of weights or channels, enabling both structured and unstructured pruning; (2) a dedicated pruning schedule $f(e)$, parameterized by the training epoch $e$ and a predefined pruning ratio $r$, to control sparsity and accommodate computational and storage constraints; and (3) systematic elimination of low-importance parameters or channels based on the ranking metric $\mathcal{M}$ to achieve efficient compression with minimal accuracy loss. These principles are implemented in Algorithms \ref{alg:FSNP} and apply to both untrained and pre-trained models.

\begin{algorithm}[htb]
	\caption{{\bf F-statistic Screening Network Pruning (FSNP)}}
	\label{alg:FSNP}
	\begin{algorithmic}
		\STATE {\bfseries Input:} Training set ${(\mathbf{x}^i, y^i)}_{i=1}^{N}$, pruning ratio $r$, global pruning schedule $f(e; r)$, and a DNN or CNN model.
		\STATE {\bfseries Output:} Pruned model with weights or channels determined by the pruning ratio $r$ in the parameter space $\mathcal{W}$ or $\mathcal{C}$.
	\end{algorithmic}
	\begin{algorithmic} [1]
		\STATE If the model is not pre-trained, first train it to a satisfactory level.
		\FOR {$e = 1$ to $E$}
			\STATE Sequentially update $\mathcal{W} \leftarrow \mathcal{W} - 
            \eta\frac{\partial L(\mathcal{W})}{\partial \mathcal{W}}$ via backpropagation.
            \STATE Update the F-statistic score $F$ using the batch data and compute the ranking metric $\mathcal{M}$.
            \STATE According to the pruning schedule, either continue training or prune the network by retaining the top $f(e; r)$ weights or channels in $\mathcal{W}$ or $\mathcal{C}$ based on the ranking metric $\mathcal{M}$.
		\ENDFOR
		\STATE Fine-tune the pruned model if needed.
\end{algorithmic}
\end{algorithm}

\begin{figure}[t]
\centering
\includegraphics[width=8.8cm]{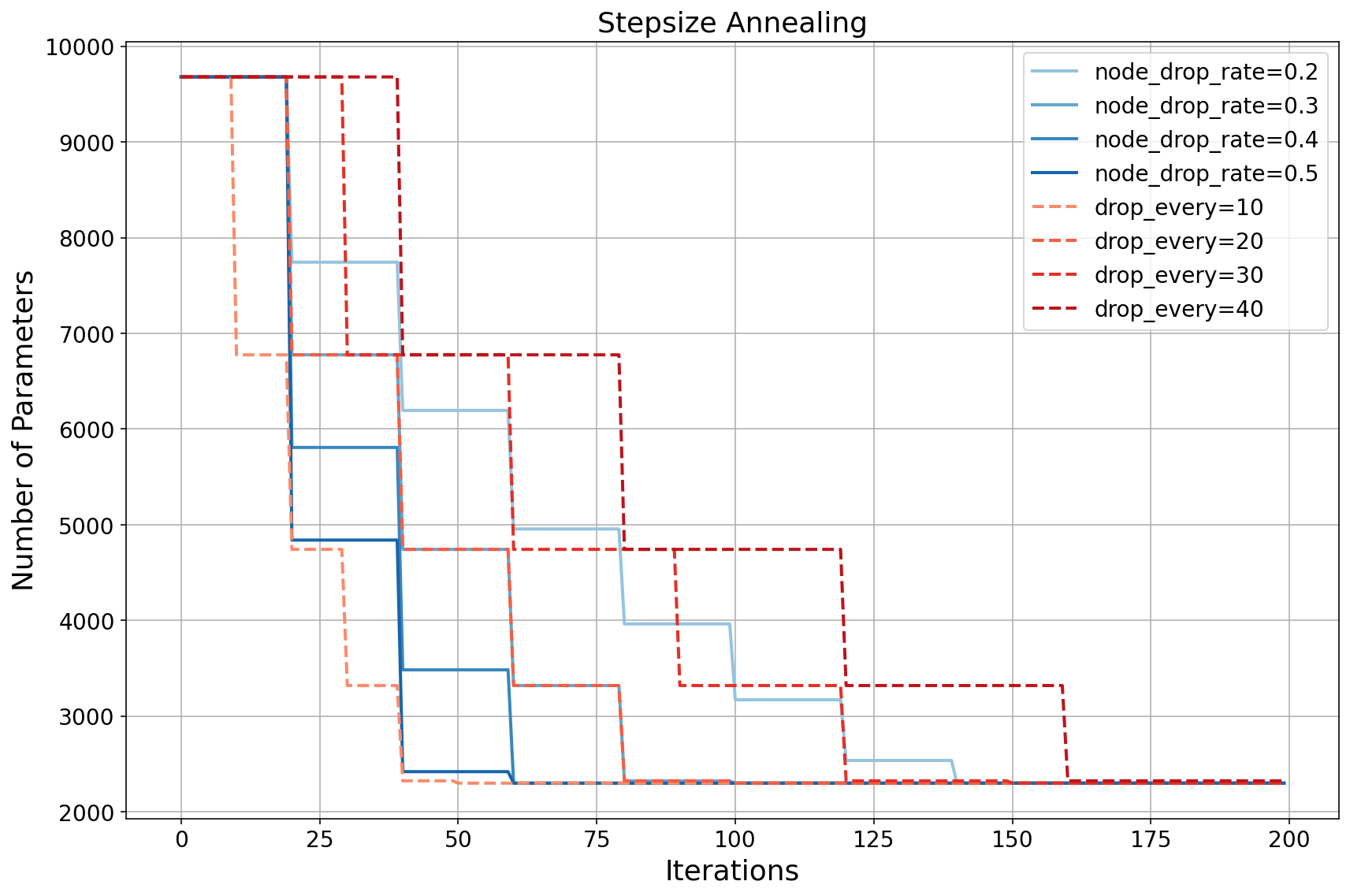}
\caption{Stepsize Annealing Schedule with Different Drop Frequency and Drop Rate}
\label{fig:Stepsize_decay}
\end{figure}

We define a step-size annealing function $f(e; r)$ to gradually adjust the pruning rate across training iterations, as illustrated in Figure \ref{fig:Stepsize_decay}. The pruning schedule is formulated as a function of the training epoch $e$ and the pruning ratio $r$. Starting from epoch $k$, pruning is applied every $d$ epochs, removing a fraction $r$ of the remaining parameters at each step. This results in a decay in the number of retained weights or channels, while ensuring that it does not fall below a predefined threshold $n_t$, where $n_0$ denotes the initial number of weights or channels in the model.

Since a global pruning ratio $r$ is employed, it may lead to degenerate cases where all weights or channels in a layer are removed. To address this issue, we introduce a per-layer minimum retention constraint. Specifically, each layer is required to retain at least $5\%–10\%$ of its original weights or channels, ensuring sufficient representational capacity and maintaining network stability during pruning.
\vspace{-1mm}
\[
f(e;r) =
\begin{cases}
    \begin{aligned}
    &n_0 & \text{if } e < k \\
    &\max\left(n_0 \cdot \left(1 - r\right)^{\left\lfloor \frac{e- k}{d} \right\rfloor}, n_t\right) & \text{if } k \leq e \leq E
    \end{aligned}
\end{cases}
\]

The FSNP algorithm prunes neural networks efficiently by iteratively or in a one-shot manner, removing the least important weights or channels, guided by a predefined pruning schedule. For fully connected layers in CNNs, each neuron is treated as an individual channel, and channel-wise pruning is applied accordingly. The decision to retain or remove a weight or channel is governed by a predefined ranking metric $\mathcal{M}$, independent of the objective loss function $L$. This independence distinguishes our method from many existing pruning approaches that require modifications to the loss function, making them less suitable for pre-trained models.

Following pruning, the model is fine-tuned to recover any potential degradation in performance.

\section{Experiments}

In this section, we first demonstrate unstructured weight pruning on the  LeNet-300-100\cite{lecun1998gradient} using the MNIST dataset \cite{lecun-mnisthandwrittendigit-2010}. Next, we conduct experiments on structured channel pruning on CifarNet \cite{krizhevsky2009learning}, ResNet \cite{he2016deep}, and DenseNet \cite{huang2017densely}, using the CIFAR\cite{krizhevsky2009learning} and SVHN\cite{netzer2011reading} datasets.

\subsection{Unstructured Weight Pruning on MNIST}

%The MNIST dataset \cite{lecun-mnisthandwrittendigit-2010} includes grayscale images of handwritten digits from 0 to 9. It contains 50K training samples, 10K validation samples, and 10K testing samples. Each image is 28x28 pixels in size. Its small size make it very popular for the testing and comparison of new algorithms in classification tasks. In this section, we will evaluate the WLS pruning method on LeNet-300-100.

We go about simultaneous training and pruning on LeNet-300-100. We use SGD to train and simultaneously prune LeNet-300-100 for $80$ epochs in total, with a minibatch size of $128$ on the MNIST dataset. The initial learning rate is $0.1$ and is divided by half every $10$ epochs. Weight decay is $1 \times 10^{-4}$ and Nesterov momentum \cite{Sutskever2013OnTI} with a value of $0.9$ (no dampening) is used to assist convergence.

\begin{table}[ht]
\centering
\begin{tabular}{llll}
\hline
\hline
Model  & Best Error & Params &Prune Rate   \\
\hline
Lenet-300-100 (Baseline)  &1.64\% &267k &- \\
Lenet-300-100 (Han \cite{han2015learning}) &1.59\% &22K &91.8\% \\
Lenet-300-100 (Barbu \cite{Guo9533741}) &1.57\% &17.4K &93.5\% \\
Lenet-300-100 (Ours) &\bf{1.51\%} &\bf{11.5K} &\bf{95.7\%} \\
\hline
\hline
\end{tabular}
\vspace{+2mm}
\caption{Unstructured weight pruning comparison on LeNet-300-100.}
\label{tab:MNIST_1}
\end{table}

\begin{table}[htb]
\centering
\begin{tabular}{llllll}
\hline
\hline
Model  & Layer & Params. & Han\% & Barbu\%     &  Ours\%  \\
\hline
                &fc1   &236K  &8\%   &4.6\%     &\textbf{4.1}\%    \\
Lenet-300-100   &fc2   &30K   &9\%   &20.1\%    &\textbf{5.3}\%   \\
                &fc3   &1K    &26\%  &68.5\%    &\textbf{22.4}\%  \\
                &Total &267K  &8.2\% &6.5\%     &\textbf{4.3}\%    \\
\hline
\hline
\end{tabular}
\vspace{+2mm}
\caption{Layer by layer compression comparisons on LeNet-300-100.}
\label{tab:MNIST_2}
\end{table}

\begin{table}[ht]
\centering
\begin{tabular}{llllll}
\hline
\hline
LeNet  & $\alpha = 0.2$ & $\alpha = 0.4$ & $\alpha = 0.6$ & $\alpha = 0.8$ & $\alpha = 1.0$  \\
\hline
Best Error  & 1.57\% & \textbf{1.51\%} & 1.69\% & 2.40\% & 6.46\% \\
\hline
\hline
\end{tabular}
\vspace{+2mm}
\caption{Best error with Ranking metric by different $\alpha$ on LeNet-300-100.}
\label{tab:MNIST_3}
\end{table}

Tables \ref{tab:MNIST_1}, \ref{tab:MNIST_2}, and \ref{tab:MNIST_3} summarize the effectiveness of the proposed unstructured screening-based pruning. Optimal performance occurs at $\alpha=0.4$, pruning 95.7\% of weights while achieving a test error of only 1.51\%. Compared to \cite{han2015learning} and \cite{Guo9533741}, the F-statistic screening approach attains higher accuracy and removes more redundant parameters. Conversely, $\alpha = 1.0$ performs worst, indicating that relying solely on the F-statistic score may not capture the true importance of each weight. Incorporating weight magnitude into the ranking metric significantly improves pruning effectiveness, achieving a better balance between compression and accuracy.

Table \ref{tab:MNIST_2} presents layer-wise compression comparisons. While overall accuracy is similar across methods, the resulting network structures differ. For LeNet-300-100, the first layer is pruned the most, suggesting that it removes redundant features early, while later layers retain enough parameters to process informative features into classes.

%These are summarized in the above tables, which highlight the effectiveness of the proposed screening-based pruning. The best performance is at $\alpha = 0.4$, with $95.7\%$ of the weights pruned at a test error of only $1.51\%$. F-statisic screening pruning approach attains higher accuracy and also eliminates more redundant parameters compared to the pruning methods of \cite{han2015learning} and \cite{Guo9533741}.

%In contrast, the poorest performance is $\alpha = 1.0$, reflecting that the reliance solely on F-statistic screening score may fail to capture true importance of each weight. While it is quite intuitive, explicit inclusion of weight magnitude into the ranking metric makes a major improvement in pruning effectiveness. As a matter of fact, this yields an optimum tradeoff between model compression and accuracy.

%Table \ref{tab:MNIST_2} lists the layer-wise comparison of achieved compression rates against state-of-the-art methods. Although all three pruning methods give roughly the same overall accuracy, they generate distinctly different network structures. In LeNet-300-100, for example, the first layer is pruned the most, that probably implies that after the first full-connected layer reduces most of the useless features in this dataset, the output fully-connected layer keeps a number of parameters big enough to process these informative features further into classes.

\subsection{Structured Channel Pruning on CIFAR and SVHN}

%The CIFAR-10 dataset \cite{krizhevsky2009learning} comprises 60K natural color images of size $32 \times 32$, including 50K training samples and 10K test samples, spanning 10 object categories with 6K images per class. The SVHN dataset \cite{netzer2011reading} consists of digit images extracted from natural scenes, featuring complex backgrounds and real-world variations. It contains approximately 600K training samples and 26K test samples, with all images resized to $32 \times 32$ pixels.

We evaluate the effectiveness of our FSNP pruning method on several deep neural network architectures. Specifically, CifarNet and ResNet-56 are evaluated on CIFAR-10, while DenseNet-40 and ResNet-164 are evaluated on SVHN. Each network architecture is adapted to the corresponding dataset and trained from scratch to establish baseline performance, consistent with prior work \cite{liu2017learning, gao2024bilevelpruning}.

Training is conducted for 160 epochs on CIFAR-10 and 80 epochs on SVHN, using a batch size of 64. All models are optimized with stochastic gradient descent (SGD), using an initial learning rate of 0.1, momentum of 0.9, and weight decay of $1 \times 10^{-4}$. The learning rate is decayed by a factor of 10 at $50\%$ and $75\%$ of the total training epochs. Standard data augmentation techniques, including normalization, random cropping, and horizontal flipping, are applied during training. After training, we perform global pruning with a predefined pruning ratio $r$, followed by fine-tuning under a step-size learning schedule. The best-performing results are reported in the following tables.

\begin{table}[ht]
\small
\centering
\begin{tabular}{lllll}
\hline
\hline
CNN &Model  & Error &\text{FLOPs}$\downarrow$ &\text{\#Params}$\downarrow$   \\
\hline
            &Baseline  &7.64\% &- &-   \\
            &FBS \cite{gao2018dynamic}   &10.12\% &74.6\%  &11.0\%   \\
CifarNet    &SEP \cite{chen2020storage}   &8.77\% &74.5\%  &\textbf{22.0\%}   \\
            &UDSP \cite{gao2024bilevelpruning}   &8.11\% &75.1\%  &20.1\%   \\
            &FSNP (Ours)      &\textbf{8.02\%} &\textbf{75.2\%}  &20.8\%   \\ 
\hline
            &Baseline   &6.88\% &- &-   \\
            &DSA \cite{ning2020dsa}          &7.09\% &\textbf{52.2\%}  &-   \\
ResNet-56    &SEP \cite{chen2020storage}          &6.56\% &50.0\%  &19.8\%   \\
            &UDSP \cite{gao2024bilevelpruning}          &6.22\% &50.0\%  &20.0\%   \\
            &FSNP (Ours)             &\textbf{6.11\%} &51.3\%  &\textbf{21.8\%}   \\ 
            
\hline
\hline
\vspace{+1mm}
\end{tabular}
\caption{Pruning performance results comparison on CIFAR-10.}
\label{tab:CIFAR10}
\end{table}

\begin{table}[ht]
\centering
\begin{tabular}{llllll}
\hline
\hline
Best Error  & $\alpha = 0.2$ & $\alpha = 0.4$ & $\alpha = 0.6$ & $\alpha = 0.8$ & $\alpha = 1.0$  \\
\hline
CifarNet  & 8.09\% & \textbf{8.02\%} & 8.61\% & 9.89\% & 11.08\% \\
\hline
ResNet-56  & \textbf{6.11\%} & 6.82\% & 7.69\% & 8.33\% & 9.42\% \\
\hline
\hline
\end{tabular}
\vspace{+2mm}
\caption{CIFAR-10: Best error with Ranking metric by different $\alpha$ on CifarNet and ResNet-56.}
\label{tab:CIFAR10_1}
\end{table}

We present CIFAR-10 results in table \ref{tab:CIFAR10}. On CifarNet, our method achieves a compelling balance between parameter efficiency and accuracy. Compared to FBS, it reduces model parameters by nearly 28\% while improving accuracy by more than 2\%, highlighting the effectiveness of our pruning strategy. UDSP achieves comparable parameter savings, yet its accuracy gain falls short of ours by 0.09\%. Even when SEP preserves all channels, our approach still surpasses it in accuracy while consuming fewer FLOPs. For ResNet-56, all pruning methods cut FLOPs by roughly 50\%. Within this context, our method attains pruning rates similar to SEP but outperforms it by 0.44\% in accuracy. While DSA achieves the highest FLOPs reduction, our approach still delivers 0.98\% higher accuracy, using comparable FLOPs overall. These results indicate that our method not only maintains desirable performance but also optimizes storage efficiency more effectively than both static and dynamic pruning strategies.

Table \ref{tab:CIFAR10_1} reports the best test errors obtained using the ranking metric across different scale values for CifarNet and ResNet-56 on CIFAR-10. The weakest pruning performance occurs at $\alpha = 1.0$, indicating that relying solely on the F-statistic screening score may not adequately capture the importance of each channel. In contrast, incorporating weight magnitude into the ranking metric enhances overall pruning effectiveness. This results in the lowest test error of $8.02\%$ for CifarNet at $\alpha = 0.4$, while ResNet-56 achieves its best test error of $6.11\%$ at $\alpha = 0.2$.

We report SVHN results in Tables \ref{tab:SVHN} and \ref{tab:SVHN_1}. On DenseNet-40, our method outperforms SLIM and DSC, achieving the lowest test error while also delivering the most effective FLOPs and parameter reduction. For ResNet-164, our approach achieves the best test error, maintaining a FLOPs reduction comparable to STNPF and slightly improving parameter reduction compared to NPRW. These results highlight the efficiency and effectiveness of our method. Consistent with the findings on CifarNet and ResNet-56 for CIFAR-10, the best pruning performance is obtained when the ranking metric combines F-score and weight magnitude with $\alpha = 0.2$ or $0.4$. Using either parameter magnitude or F-score alone does not achieve optimal results, underscoring the benefit of the hybrid ranking strategy.

\begin{table}[ht]
\small
\centering
\begin{tabular}{lllll}
\hline
\hline
CNN &Model  & Error &\text{FLOPs}$\downarrow$ &\text{\#Params}$\downarrow$   \\
\hline
                &Baseline  &1.89\% &- &-   \\
DenseNet-40      &SLIM \cite{liu2017learning}   &1.81\% &49.8\%  &56.6\%   \\
                &DSC \cite{Guo9533741}   &1.80\% &45.1\%  &54.9\%   \\
                &FSNP (Ours)      &\textbf{1.78\%} &\textbf{50.2\%}  &\textbf{56.8\%}   \\ 
\hline
                &Baseline  &1.78\% &- &-   \\
                &SLIM \cite{liu2017learning}   &1.81\% &54.9\%  &34.3\%   \\
ResNet-164       &STNPF \cite{liu2022evolving}   &1.74\% &\textbf{63.0\%}  &-   \\
                &NPRW \cite{pei2022neural}   &3.37\% &58.9\%  &37.1\%   \\             
                &FSNP (Ours)      &\textbf{1.71\%} &59.1\%  &\textbf{37.4\%}   \\ 

\hline
\hline
\vspace{+1mm}
\end{tabular}
\caption{Pruning performance results comparison on SVHN.}
\label{tab:SVHN}
\end{table}

\begin{table}[ht]
\centering
\begin{tabular}{llllll}
\hline
\hline
Best Error  & $\alpha = 0.2$ & $\alpha = 0.4$ & $\alpha = 0.6$ & $\alpha = 0.8$ & $\alpha = 1.0$  \\
\hline
DensetNet-40  & \textbf{1.78\%} & 1.79\% & 2.41\% & 3.89\% & 5.08\% \\
\hline
ResNet-164  & 1.77\% & \textbf{1.71\%} & 2.99\% & 3.48\% & 5.82\% \\
\hline
\hline
\end{tabular}
\vspace{+2mm}
\caption{SVHN: Best error with Ranking metric by different $\alpha$ on DenseNet40.}
\label{tab:SVHN_1}
\end{table}

\subsection{Ablation Study}
This section analyzes the design choices and hyperparameter settings of our proposed pruning framework.

\begin{figure*}[htb]
\centering
\begin{tabular}{cc}
\includegraphics[width=7.8cm]{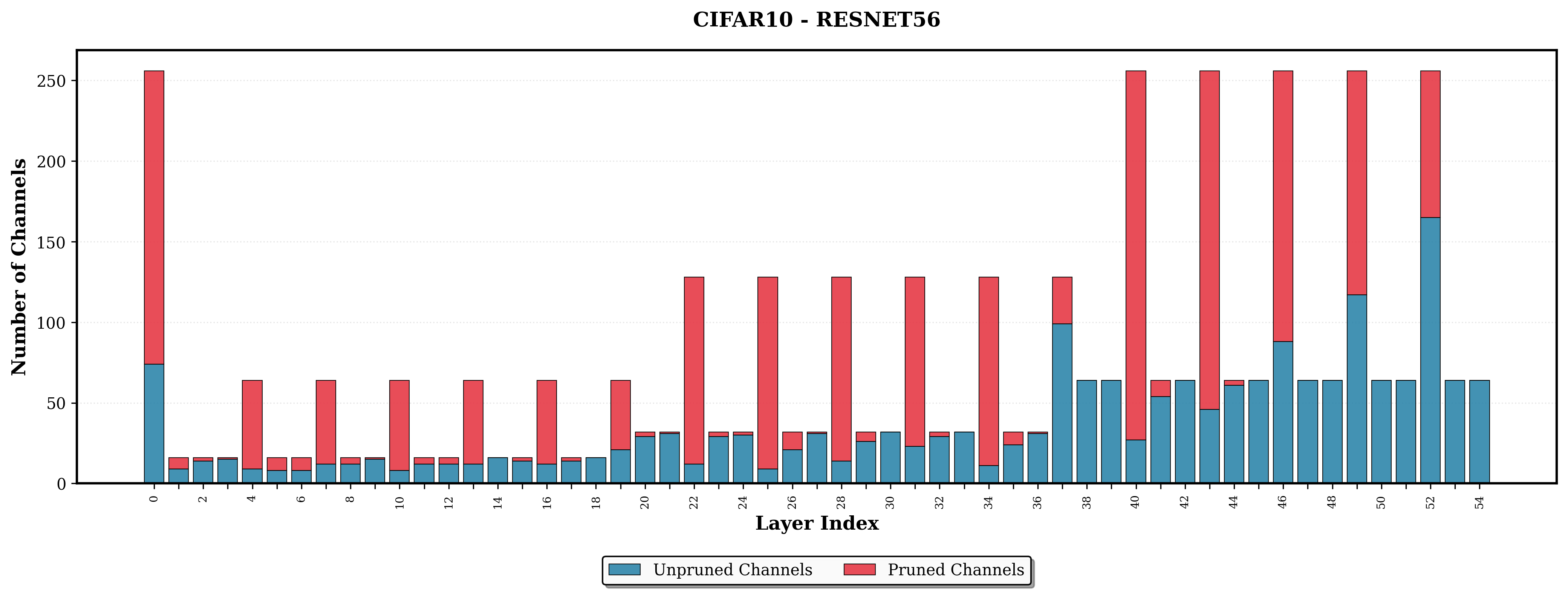}
&\includegraphics[width=7.8cm]{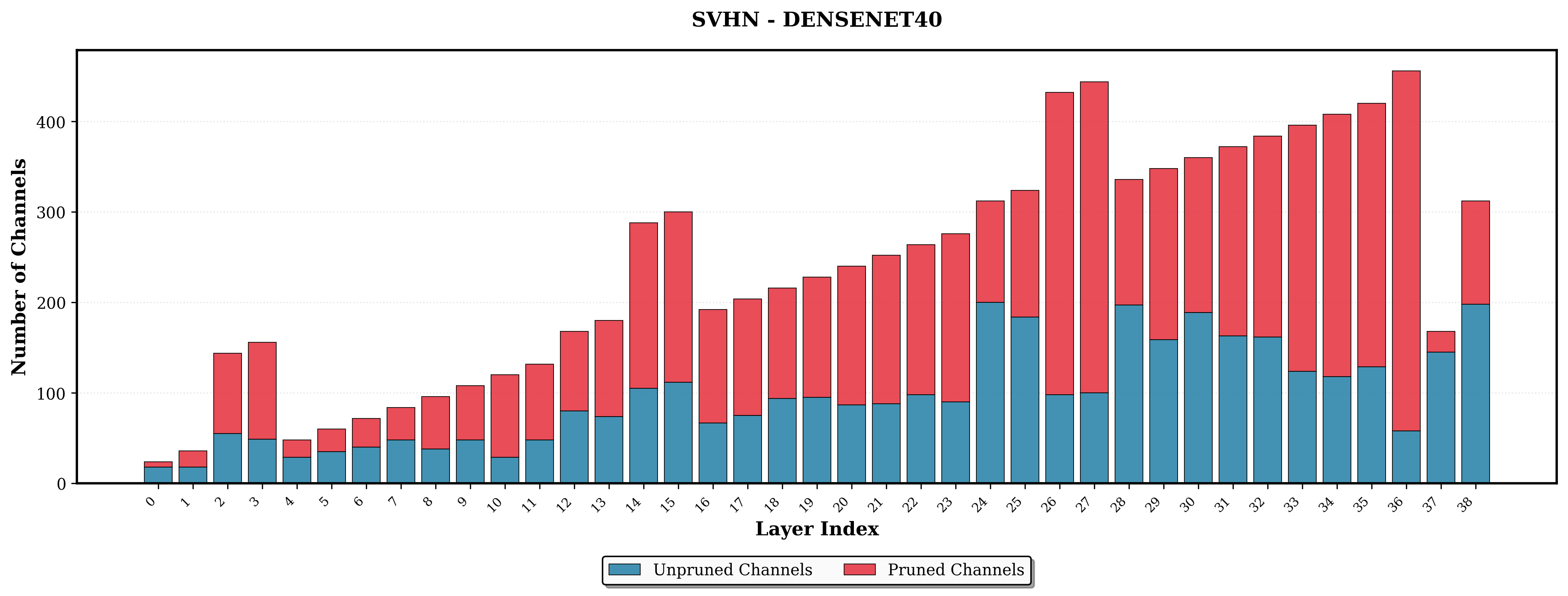}
\end{tabular}
\caption{Original and Remaining Channel Counts after Pruning (Left: ResNet-56 on CIFAR-10; Right: DenseNet-40 on SVHN).}\label{fig:channel_distribution}
\end{figure*} 

\begin{figure*}[htb]
\centering
\begin{tabular}{cc}
\includegraphics[width=7.8cm]{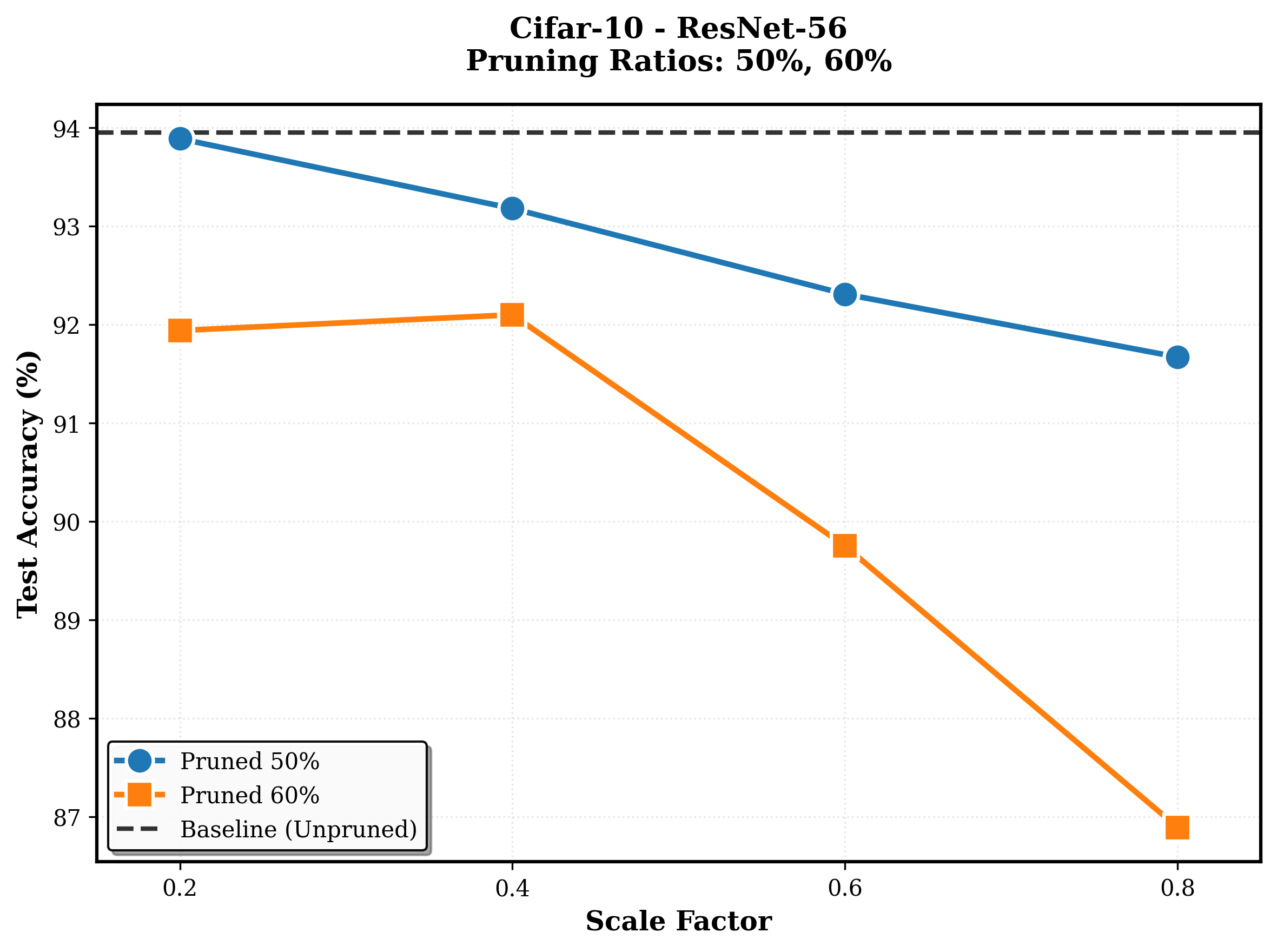}
&\includegraphics[width=7.8cm]{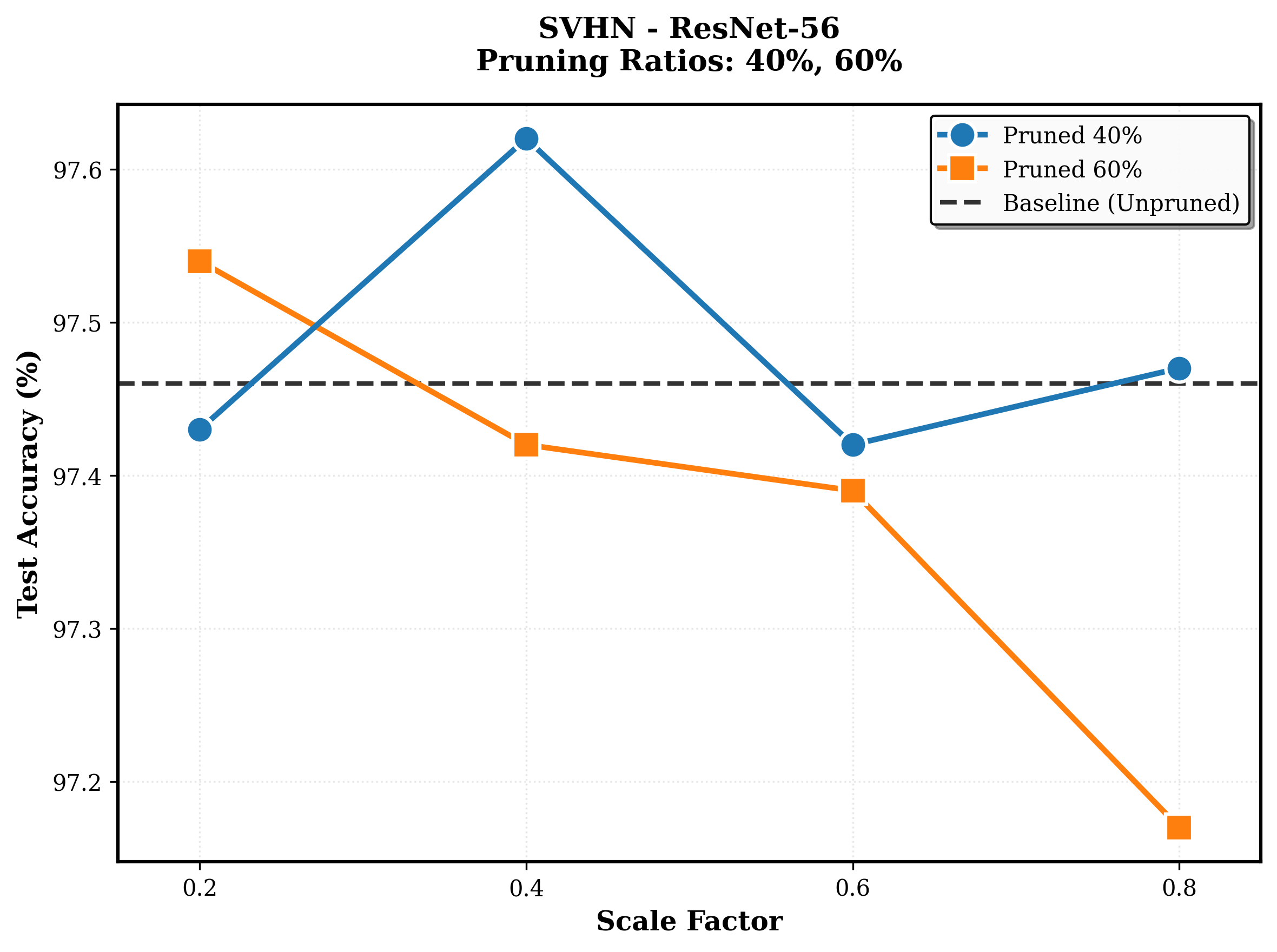}
\end{tabular}
\caption{Test Error vs. Scale Value under Different Channel Pruning Ratios on ResNet-56
(Left: 50\% and 60\% pruning on CIFAR-10; Right: 40\% and 60\% pruning on SVHN).}\label{fig:scale_value_range}
\end{figure*} 

We illustrate the impact of channel pruning on ResNet-56 and DenseNet-40 by presenting histograms of the remaining channel ratios in Fig. \ref{fig:channel_distribution} for ResNet-56 on CIFAR-10 and DenseNet-40 on SVHN, both with $\alpha = 0.2$. Due to the substantial architectural differences between the two networks, the resulting channel distribution patterns exhibit clear and distinct characteristics.

For ResNet-56, layers with a large number of channels undergo more aggressive pruning, retaining only a small fraction of their original channels, while layers with fewer channels preserve a much higher proportion. This behavior is intuitive: convolutional layers with higher channel capacity are more likely to contain redundant or less informative channels. Moreover, pruning in ResNet-56 is not only concentrated in high-channel layers but also predominantly occurs in the earlier stages of the network. Layers beyond stage 38, which have relatively small channel widths, are barely pruned. This uneven pruning pattern suggests that ResNet-56 is layer-wise over-parameterized for the CIFAR-10 task, particularly in its early and high-capacity layers, and that many channels in these layers are unnecessary for maintaining performance.

In contrast, DenseNet-40 with a growth rate of 12 exhibits a more uniform distribution of retained channels across most convolutional layers. This observation is well aligned with the DenseNet architecture: each dense block receives feature maps from all preceding blocks, leading to smoother and more evenly distributed activation channels throughout the network. As a result, pruning in DenseNet-40 affects nearly every layer to a similar extent, rather than being concentrated in early layers as in ResNet-56. This structural characteristic enables DenseNet to distribute representational capacity more evenly across layers, resulting in a more balanced pruning behavior.

Fig \ref{fig:scale_value_range} illustrates the relationship between test error and scale value under different channel pruning ratios for ResNet-56, with results on CIFAR-10 (50\% and 60\% pruning) shown on the left and results on SVHN (40\% and 60\% pruning) shown on the right. This visualization is used to analyze how different scale values in the ranking metric affect the final pruning performance across multiple pruning ratios.

As shown in the Fig \ref{fig:scale_value_range}, when the scale value exceeds $0.6$, ResNet-56 consistently exhibits degraded test accuracy after channel pruning. In contrast, the lowest test errors are achieved when the scale value is relatively small (e.g., $0.2$ or $0.4$), where the ranking metric effectively balances the F-statistic screening score and channel importance. These results are consistent with the observations in the previous section, indicating that optimal pruning performance is obtained when the ranking metric combines F-score and weight magnitude with a small $\alpha$. Relying solely on either parameter magnitude or F-score fails to produce the best results, further highlighting the advantage of the proposed hybrid ranking strategy.

\begin{figure*}[htb]
\centering
\begin{tabular}{cc}
\includegraphics[width=8.2cm]{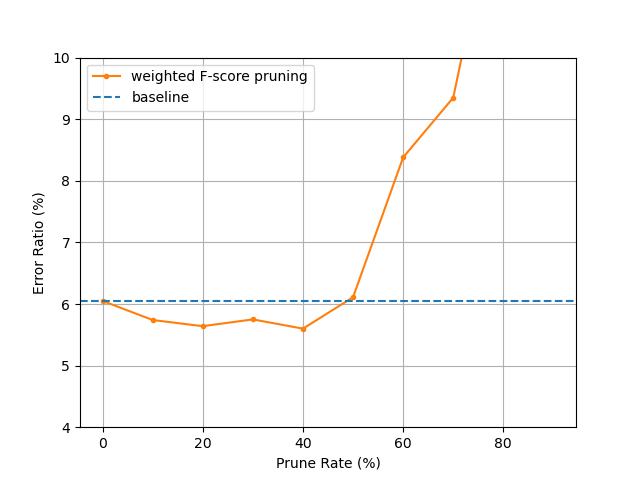}
&\includegraphics[width=8.0cm]{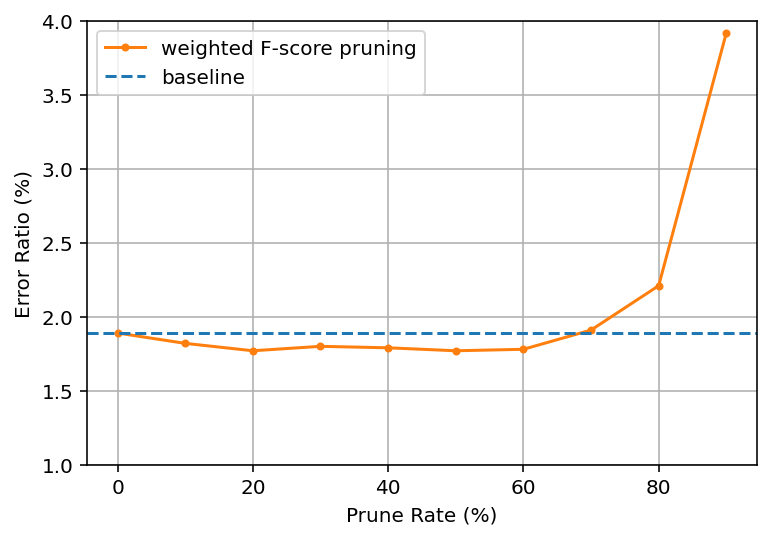}
\\ ResNet-56 on CIFAR-10
& DenseNet-40 on SVHN
\end{tabular}
\caption{Test Error vs. Channel Prune Ratio (Left: ResNet-56 on CIFAR-10; Right: DenseNet-40 on SVHN).}\label{fig:prune_ratio_range}
\end{figure*} 

Selecting an appropriate channel pruning ratio is a crucial design decision. Pruning too few channels yields limited computational and memory savings, whereas overly aggressive pruning can severely damage the network, preventing fine-tuning from fully recovering performance. To examine this trade-off, Fig. \ref{fig:prune_ratio_range} shows the relationship between test error and pruning ratio for ResNet-56 on CIFAR-10 and DenseNet-40 on SVHN, with $\alpha = 0.2$. The results indicate that model accuracy remains largely unaffected as the pruning ratio increases, up to a critical threshold. Within this regime, fine-tuning is generally sufficient to compensate for the minor performance degradation introduced by pruning. Beyond this point, however, the representational capacity of the network becomes insufficient. When the pruning ratio exceeds approximately 80\%, even fine-tuned models begin to underperform relative to the unpruned baseline. This behavior is consistent with prior observations \cite{liu2017learning}, highlighting the existence of a practical upper bound on pruning aggressiveness for maintaining model accuracy.

\section{Conclusion}

Deep neural networks have brought about a revolution in machine learning and, today, form a very strong machinery of learning complex representations from diverse domains. Despite their great success, the modern DNNs also suffer from severe computational challenges. The training and inference of modern DNNs become extremely resource-intensive due to tens to hundreds of millions of parameters. In this paper, a network-pruning technique has been presented that removes unnecessary parameters to alleviate these issues. Pruning has been done based on the statistical impact of every network parameter on classification categories. Contribution analysis for the connections and channels using screening methods in our proposed approach includes unstructured and structured pruning, which gives the leverage of removing unnecessary elements without affecting model performance. Our extensive experimental evaluation using real-world vision datasets, including both FNNs and CNNs, shows that the proposed screening approach alone could give promising results but cannot always outperform state-of-the-art pruning methods. In contrast, the hybrid method with the magnitude-based pruning approach yields much better or at least highly competitive performance, thereby presenting a practical solution toward neural network optimization in efficiency.

\bibliographystyle{IEEEtran}
\bibliography{refs}

\end{document}